\documentclass[letterpaper, 10 pt, conference]{ieeeconf}
\IEEEoverridecommandlockouts    
\usepackage{graphics}           
\usepackage{times}              
\usepackage{amsmath}            
\usepackage{amssymb}            
\usepackage{graphicx}
\usepackage{algorithm}
\usepackage[noend]{algpseudocode}
\usepackage{booktabs}
\usepackage{color}
\usepackage{subfigure}
\usepackage{multirow}
\usepackage{setspace}
\definecolor{instructioncolor}{rgb}{.5,.5,.5}
\usepackage{threeparttable}

\makeatletter
\renewcommand{\maketag@@@}[1]{\hbox{\m@th\normalsize\normalfont#1}}%
\makeatother

\usepackage[font=small]{caption}

\def\secref#1{Sec.~\ref{#1}}
\def\figref#1{Fig.~\ref{#1}}
\def\tabref#1{Tab.~\ref{#1}}
\def\eqref#1{Eq.~(\ref{#1})}
\def\algref#1{Alg.~\ref{#1}}

\makeatletter
\usepackage{xspace}
\DeclareRobustCommand\onedot{\futurelet\@let@token\@onedot}
\def\@onedot{\ifx\@let@token.\else.\null\fi\xspace}

\def\etal{{et al}\onedot}
\makeatother

\def\etalcite#1{\etal~\cite{#1}}

\usepackage{array}
\newcolumntype{L}[1]{>{\raggedright\let\newline\\\arraybackslash\hspace{0pt}}m{#1}}
\newcolumntype{C}[1]{>{\centering\let\newline\\\arraybackslash\hspace{0pt}}m{#1}}
\newcolumntype{R}[1]{>{\raggedleft\let\newline\\\arraybackslash\hspace{0pt}}m{#1}}

\newcommand{\RR}{\mathbb{R}}

\renewcommand{\b}[1]{\mbox{\boldmath$#1$}}

\newcommand{\m}[1]{{\mbox{{\sffamily\slshape{#1\/}}}}}

\newcommand{\tr}[0]{\sf T}              

\newcommand{\bB}{\b B}
\newcommand{\bC}{\b C}

\newcommand{\bF}{\b F}

\newcommand{\bI}{\b I}

\newcommand{\bM}{\b M}

\newcommand{\bO}{\b O}

\newcommand{\bQ}{\b Q}

\newcommand{\bT}{\b T}

\newcommand{\bp}{\b p}

\newcommand{\mS}{\m S}

\usepackage{url}

\title{\LARGE \bf InsMOS: Instance-Aware Moving Object Segmentation in LiDAR Data}

\author{Neng Wang \and Chenghao Shi \and Ruibin Guo \and Huimin Lu$^*$ \and Zhiqiang Zheng\and Xieyuanli Chen$^*$
  \thanks{All authors are with the College of Intelligence Science and Technology, National University of Defense Technology, Changsha, China.}%
  \thanks{$^*$corresponding authors, \{lhmnew, xieyuanli.chen\}@nudt.edu.cn}
\thanks{This work was supported in part by the National Science Foundation of China under Grant U1913202 and U22A200600, as well as Major Project of Natural Science Foundation of Hunan Province under Grant 2021JC0004.
}%
}

\begin{document}
\maketitle
\thispagestyle{empty}
\pagestyle{empty}

\begin{abstract}
	Identifying moving objects is a crucial capability for autonomous navigation, consistent map generation, and future trajectory prediction of objects. In this paper, we propose a novel network that addresses the challenge of segmenting moving objects in 3D LiDAR scans. Our approach not only predicts point-wise moving labels but also detects instance information of main traffic participants. Such a design helps determine which instances are actually moving and which ones are temporarily static in the current scene. Our method exploits a sequence of point clouds as input and quantifies them into 4D voxels. We use 4D sparse convolutions to extract motion features from the 4D voxels and inject them into the current scan. Then, we extract spatio-temporal features from the current scan for instance detection and feature fusion. Finally, we design an upsample fusion module to output point-wise labels by fusing the spatio-temporal features and predicted instance information. We evaluated our approach on the LiDAR-MOS benchmark based on SemanticKITTI and achieved better moving object segmentation performance compared to state-of-the-art methods, demonstrating the effectiveness of our approach in integrating instance information for moving object segmentation. Furthermore, our method shows superior performance on the Apollo dataset with a pre-trained model on SemanticKITTI, indicating that our method generalizes well in different scenes.
	The code and pre-trained models of our method will be released at \url{https://github.com/nubot-nudt/InsMOS}.
\end{abstract}

\section{Introduction}
\label{sec:intro}

3D LiDAR-based environment perception is important for autonomous navigation systems due to its robustness to illuminational changes and large field of view~\cite{kim2022ral}. 
However, moving objects can often disrupt LiDAR perception, leading to suboptimal performance in downstream tasks, such as point cloud registration~\cite{lu2019iccv}, simultaneous localization and mapping (SLAM)~\cite{chen2019iros,chen2020iros}, static map creation \cite{arora2021ecmr,kim2020iros}, and path planning \cite{chen2022neuroc,lee2023ral}. 
Therefore, identifying moving objects in LiDAR data is one of the most critical abilities for LiDAR-based applications of autonomous mobile systems.

\begin{figure}[ht]
	\subfigure[4DMOS]{
		\hspace{-0.2cm}
			\includegraphics[width=0.98\linewidth]{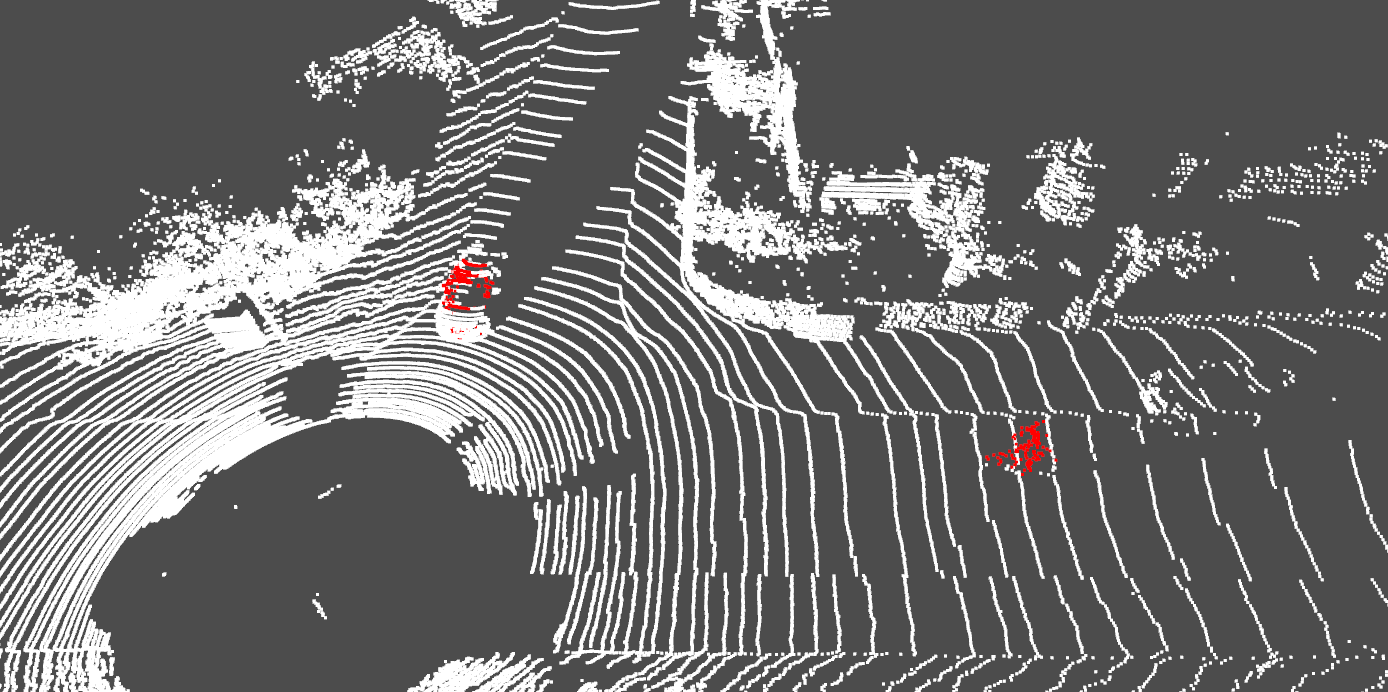}

	}
    \vspace{-0.2cm}
	\quad
    
	\subfigure[Ours]{
		\hspace{-0.2cm}
		\includegraphics[width=0.98\linewidth]{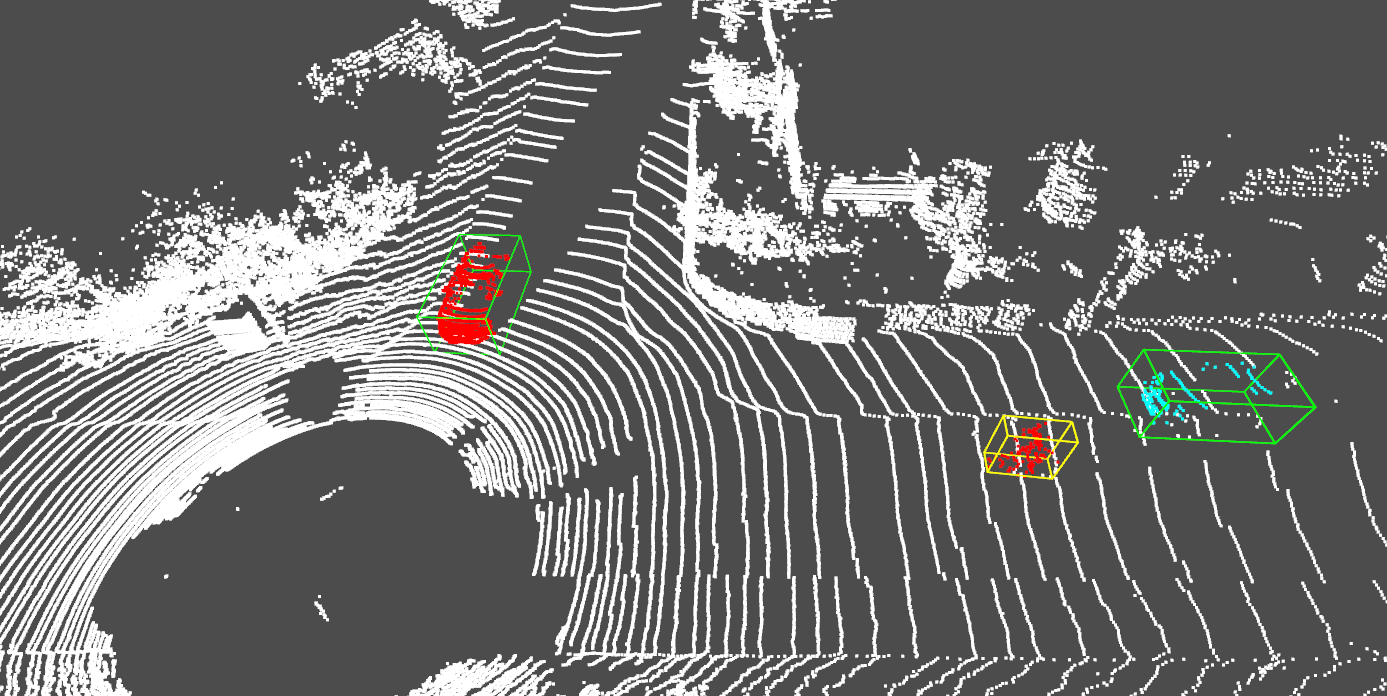}

    }
	\caption{
		The comparison of moving object segmentation between our approach and 4DMOS. (a) 4DMOS only segments part of the LiDAR points of a moving object due to the lack of instance information. (b) Our approach detects all movable objects and distinguishes their category, displayed with different colored bounding boxes. By utilizing the instance information, we can completely segment a moving instance (red) and determine which instance is currently static (cyan).
	}
	\label{fig:motivation}
	\vspace{-0.5cm}
\end{figure}

In this work, we aim to tackle the challenge of segmenting moving objects from static environments using consecutive LiDAR point clouds. 
In order to achieve moving object segmentation (MOS) in LiDAR data, it is usually necessary to extract temporal features from the sequence of 3D LiDAR scans. Such features are used to determine whether object's position has changed compared to the previous observations.
Many existing methods apply 2D convolutions to extract temporal features from residual range images~\cite{kim2022ral,chen2021ral,sun2022iros} generated by spherical projection. These methods are usually fast in operation but relatively generalizing poorly in unseen environments.
Some non-projection-based methods~\cite{mersch2022ral,kreutz2023wacv} are also able to extract temporal features directly from the sequence of point clouds.
Besides, other approaches in different domains can be used to tackle MOS task, such as scene flow estimation~\cite{baur2021iccv,liu2019cvpr,dong2022cvpr} and map cleaning~\cite{arora2021ecmr,kim2020iros,pomerleau2014icra,pagad2020icra}.

Different from existing approaches that classify moving objects on a point-by-point basis using raw LiDAR data, our method focuses more on a higher instance level and decides which instances are currently moving in the scene.
As illustrated in~\figref{fig:motivation}, our approach predicts both point-wise moving labels and instance information within the current point cloud, using a top-down and bottom-up fusion fashion, which is more natural and can generalize to different setups.
Moreover, this instance information can also be helpful for subsequent tasks, such as path planning~\cite{chen2022neuroc,lee2023ral} and future state predictions~\cite{ghorai2022tits}.

The main contribution of this paper is a novel network that can efficiently and accurately segment moving objects in 3D LiDAR data while detecting instance information of common movable objects, such as pedestrians and vehicles.
We use 4D sparse convolutions~\cite{choy2019cvpr} to extract motion features from 4D voxels created by quantifying the consecutive input scans and inject the motion features into the current scan.
Then, we concatenate the motion features with the original point features and use an instance detection module to extract spatio-temporal features from the current scan for instance detection and feature fusion.
Finally, an upsample fusion module is designed to fuse the spatio-temporal features and instance information for achieving better segmentation within the instances.
To further improve the performance, we propose an instance-based refinement algorithm to refine the network predictions and determine which instances are actually moving.
Based on instance-level understanding, our method achieves state-of-the-art performance on SemanticKITTI dataset and generalizes well to Apollo dataset without any fine-tuning.

In sum, we make three key claims:
Our approach is able to
(i) predict instances and segment moving objects simultaneously;
(ii) achieve the state-of-the-art MOS performance compared with existing methods;
(iii) generalize well to different scenes without any fine-tuning.
These claims are supported by our experimental evaluation.

\section{Related Work}
\label{sec:related}

Existing MOS methods for 3D LiDAR data can be divided into two groups, projection-based and non-projection-based approaches.

\textbf{Projection-based approaches.} 
Many existing methods convert the 3D raw point cloud into a 2D image plane, such as range images \cite{kim2022ral,chen2021ral,sun2022iros,chen2022ral} and Bird's Eye View (BEV) images \cite{baur2021iccv,mohapatra2021arxiv,wu2020cvpr} to facilitate online LiDAR point cloud processing.
To extract temporal features, they usually need to pre-process the past consecutive projected image representations.
Chen~\etalcite{chen2021ral} first exploit the residual range images for online MOS. Their proposed method gets rid of the restraint of the pre-built maps and can be directly applied in a SLAM pipeline.
Based on that, Sun~\etalcite{sun2022iros} design a dual-branch structure to extract spatial and temporal features separately, and then fuse them with motion-guided attention modules. By applying a coarse-to-fine network architecture, artifacts generated in the back-projection can be effectively reduced.
Unlike~\cite{chen2021ral,sun2022iros}, Kim~\etalcite{kim2022ral} argue that semantic information is helpful for MOS task, so they insert semantic cues into the network and achieve better segmentation performance.
Different from the spherical projection, BEV images are usually generated by compressing 3D point clouds in the height direction, which is more intuitive for observing the movement of objects.
Mohapatra~\etalcite{mohapatra2021arxiv} propose a lightweight network for processing BEV images and achieving real-time MOS on an embedded platform. However, although their method performs fast, they can only segment moving objects in BEV space with relatively low accuracy.
Wu~\etalcite{wu2020cvpr} represent a given sequence of 3D LiDAR scans into BEV maps and then extract spatio-temporal features from the maps to estimate the motion status of objects. They can not only segment moving objects at the current moment but also predict the future trajectories. However, their method can only be performed in BEV space and cannot provide point-wise predictions.
In general, such projection-based methods suffer from the loss of information and fixed pattern of the trained data, thus usually unable to generalize well into new setups and environments.

\textbf{Non-projection-based approaches.}
In order to improve the accuracy and generalization ability, some methods tackle the MOS task by directly processing 3D point clouds.
Mersch~\etalcite{mersch2022ral} propose 4DMOS that exploits sparse 4D convolutions to extract spatio-temporal features from the past consecutive point clouds and predict point-wise moving confidence scores. Their approach achieves a fine balance between performance and efficiency.
Following the 4DMOS, Kreutz~\etalcite{kreutz2023wacv} design a 4D LiDAR representation to identify moving objects by self-supervised learning, but it only works in stationary settings. 

In addition to the above mentioned, some other methods can also segment moving objects in 3D LiDAR scans, including 3D scene flow estimation~\cite{baur2021iccv,liu2019cvpr,dong2022cvpr} and map cleaning~\cite{arora2021ecmr,kim2020iros,pomerleau2014icra,pagad2020icra,lim2021ral}. 
The main purpose of 3D scene flow estimation is to predict a 3D displacement for each point between two consecutive input point clouds, which can be used to determine whether a point is moving. However, their performance on the MOS task is often restricted due to the short input time horizon.
Identifying moving objects through map cleanup requires first building a scene map, e.g., Pomerleau~\etalcite{pomerleau2014icra} maintain a global map, and segment moving points and static points based on repeated observations. Similarly, Pagad~\etalcite{pagad2020icra} create an occupancy map for detecting and removing moving objects in LiDAR scans. Most recently, Lim~\etalcite{lim2021ral} propose ERASOR achieving good map cleaning results. However, these methods are time-consuming and usually run offline.

Unlike previous approaches, we insert instance information into our network to completely segment a moving object.
We extract motion features from the sequence point clouds and inject them into each point in the current scan.
And then we extract spatio-temporal features from the current scan for instance detection and feature fusion.
Finally, we design an upsample fusion module to integrate spatio-temporal features and instance information and output point-wise moving labels of the current scan.
Besides, we also propose an instance-based refinement to further refine the predictions of our network.
To the best of our knowledge, our method is the first work to explicitly exploit instance information for LiDAR MOS task.

\section{Our Approach}
\label{sec:main}

\begin{figure*}[ht]
	\centering
	\includegraphics[width=1\linewidth]{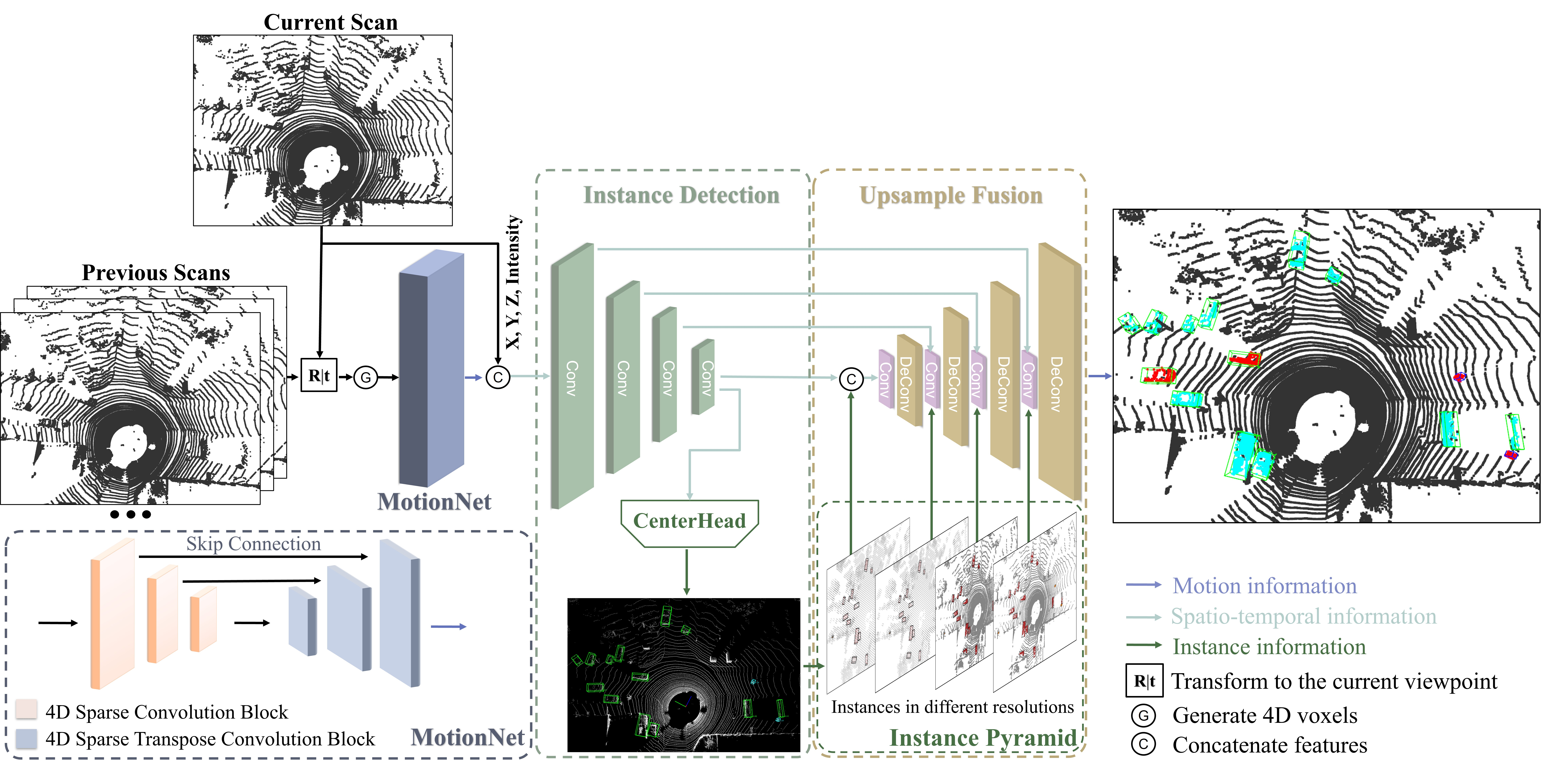}
	\caption{Overview of our network. MotionNet mainly extracts motion features. Instance Detection Module extracts spatio-temporal features and detects instances. Upsample Fusion Module is applied to fuse the spatio-temporal and instance features, and predict point-wise moving confidence scores. By inserting instance information, our approach achieves better moving object segmentation performance and can distinguish between actually moving instances (red) and static (cyan). 
	}
	\label{fig:pipeline}
\end{figure*}

\subsection{Preliminaries}
\label{sec:pre}
Given the current LiDAR scan $\mS_{0}=\{\bp_{i} \in \RR^{4}\}^{M}_{i=1}, \bp_{i}=[x_{i},y_{i},z_{i},1]^{\tr}$ with $M$ points represented as homogeneous coordinates and the past $N-1$ consecutive scans $\mS_{1},\mS_{2},...,\mS_{N-1}$, we aim to segment the moving objects in $\mS_{0}$ by classifying each point as either moving or static.
To achieve that, we utilize the spatio-temporal information of the consecutive scans.
We first align all the scans to the current viewpoint.
Assuming the relative pose transformations between past consecutive scans are provided by an odometry method, noted as $\bT_{1}^{0},\bT_{2}^{1},...,\bT_{N}^{N-1}$, we transform the past $N-1$ scans to the current scan by:
\begin{equation}
\mS_{j\rightarrow0}=\{\bp_{i}'=\bT_{j}^{0}\bp_{i}|\bp_{i}\in \mS_{j}\},~\bT_{j}^{0}\hspace{-1mm}=\hspace{-1mm}\prod_{k=0}^{j-1}\bT_{j-k}^{j-k-1}.
\end{equation}

We add an additional time dimension for each point in the aligned point clouds to feed the temporal information into our network.
As an example, for a point $\bp_{i}'$ in scan $\mS_{j\rightarrow0}$, we assign the scan time interval $t_{i}$ to the point and get $\tilde{\bp_{i}}'\hspace{-0.5mm}=\hspace{-0.5mm}[x_{i}',y_{i}',z_{i}',t_{i}]^{\tr}$. we ignore the homogeneous term for simplicity.
For memory saving and system efficiency, we quantize the aligned point clouds to 4D voxels with a time resolution $\Delta t$ and a space resolution $\Delta s$, and only use the non-empty parts as the network input.

\subsection{Network Structure}
\label{sec:net}
The overview of our network is shown in \figref{fig:pipeline}. Our network is composed of three main components: MotionNet, instance detection module, and upsample fusion module.
MotionNet is designed to extract motion features of the input 4D voxels. 
We then concatenate motion features with original point features and use an instance detection module to extract the spatio-temporal features from the current scan for instance detection and feature fusion.
Finally, the upsample fusion module achieves point-wise MOS by integrating spatio-temporal and instance information.

\subsubsection{MotionNet}
\label{sec:motionnet}
MotionNet utilizes consecutive LiDAR scans to extract motion information.
It takes the 4D voxels introduced in \secref{sec:pre} as input and outputs point-wise motion features $\bF\hspace{-0.5mm}=\hspace{-0.5mm}\{\b{f}_i\hspace{-0.5mm}\in\hspace{-0.5mm} \RR^{3}\}_{i=1}^M$ of the current scan.
We design our MotionNet based on 4DMOS~\cite{mersch2022ral}.
It utilizes a 4D sparse convolution mechanism, i.e., Minkowski engine~\cite{choy2019cvpr} and an hourglass structure to capture the features from different layers of the network. 
Furthermore, the skip connection is used to fuse the features of multiple layers of the network to maintain more details. 
Unlike 4DMOS, our MotionNet has fewer layers since we use this backbone only to extract motion features instead of directly outputting point-wise moving segmentation labels, which lightens the feature extraction backbone and is the key to achieving online performance. Even though the backbone is lightweight, our method still performs well in motion detection due to our instance-aware design detailed in the following subsections.

\subsubsection{Instance Detection Module}
Detecting moving objects through instance information is inspired by human perception of the world. As humans, we typically do not directly judge whether a point is moving or not. Instead, we first determine whether an object is moving and then reason the points on it are also moving.
This top-down reasoning can avoid a problem in existing methods, where part of the same object is classified as moving while the other is considered static.  
Based on the fact that all points on the same object should be labeled consistently, we propose an instance detection module to extract the instance features and help the following moving objects reasoning. 
In this instance detection module, we concatenate the learned point-wise motion features $\b{f}$ with point coordinates $x,y,z$, and intensity $r$ as input, which will be used to generate spatio-temporal features.
Afterwards, we use multiple layers of 3D sparse convolution~\cite{graham2018cvpr} to extract spatio-temporal features from the current scan and gradually compress it into a BEV image.
After simple 2D convolutions, CenterHead~\cite{yin2021cvpr} backbone is then used to generate $C$ heatmaps $\b{\hat{E}}$ of size $H \times W$ and instance information $\b{\hat{O}}\hspace{-1mm}=\hspace{-1mm}\{\b{\hat{o}}_u=[\tilde{x}_u,\tilde{y}_u,z_u,l_u,m_u,q_u,\text{sin}(\omega_u),\text{cos}(\omega_u)] \hspace{-1mm}\in \hspace{-1mm}\RR^8\}_{u=1}^{U}$, where $C$ represents the number of instance categories, $U$ represents the number of instances, $\tilde{x}$ and $\tilde{y}$ represent coordinate quantization offsets, $z$ represents the height, $l$, $m$ and $q$ represent 3D bounding box size and $\omega$ represents orientation.
Each score in heatmap $\hat{E}_{cij}$ represents the confidence at location $(i,j)$ belonging to the instance class $c$.
For more details about CenterHead, please refer to~\cite{yin2021cvpr}. 

\subsubsection{Upsample Fusion Module}
\label{sec:upsample}\
In upsample fusion module, we aim to resume point-level features by fusing the spatio-temporal features and the instance features.
We directly concatenate the spatio-temporal features and the instance features as the input. We then fuse them by performing convolutions, and utilize the same number of layers of 3D sparse deconvolutions~\cite{graham2018cvpr} as those convolutions used in the instance detection module to resume the point-level features hierarchically.
To strengthen the instance information in the final features and maintain more details learned in different layers, we propagate the instance features from multi-resolution point clouds and form the instance pyramid. After concatenating with the corresponding spatio-temporal features, we inject them into the upsample fusion module.

Finally, the upsample fusion module outputs the point-wise features $\bF'\hspace{-1.5mm}=\hspace{-1.5mm}\{\b{f}'_i\hspace{-1.5mm}\in \hspace{-1.5mm}\RR^3\}_{i=1}^M$, which are then used to determine the moving label of LiDAR points after passing a softmax function.
Specifically, three channels of $\b{f}'_i$ represent the confidence of a point belonging to three categories, respectively, which are unlabeled, static, and moving.
The category with the highest score is the final label of the point. 

\subsection{Moving Instance Refinement}
\label{sec:post_processing}
The network can determine the moving instances by directly checking the predicted MOS label of the instance center point. However, such pure top-down fashion may misdetect, thus enlarging the amount of misclassified points. 
To further improve the performance, we then check again the point-wise predictions within an instance and propose the instance-based refinement algorithm in a button-up fashion, as described in~\algref{alg:algorithm}.

\begin{algorithm}[t]
	\small
	\caption{ Instance-based refinement algorithm}
	\label{alg:algorithm}
	\begin{algorithmic}[1]
		\Require Moving labels ${\b{ L}}$ and confidence scores ${\b{ P}}$, and instance information ${\b{ B}}$ of network prediction, and moving label of instances $\bI$, relative pose transformations $\b{ T}$, LiDAR scan $\m S$, and other thresholds $\alpha_0$, $\alpha_1$, $\beta_0$, $\beta_1$, $\theta_{0}$, $\theta_{1}$.
		
		\Ensure	Moving label ${\b{ L}_0}'$ of each point.

		\State ${\b{ L}_0}' = {\b{ L}_0}$  
		\State $\bI_0=\text{Zeros(Length($\bB_0$))}$ 
		\State $idx\hspace{-0.0mm} = \hspace{-0.0mm}\text{Find\_index}(\mS_0,\bB_0)$ \%The index of points in instance.\hspace{0.004cm} \,
		\State $m_c = 0$  \hspace{2.475cm}\%The number of moving vehicles. 
		\State $\bQ \hspace{-0.8mm}= \hspace{-0.8mm}\text{Zeros(Length($\bB_0$))}$ \hspace{0.46cm}\%\,Instances' status in past scans.
		\State $\bM \hspace{-0.8mm}= \hspace{-0.8mm}\text{Zeros(Length($\bB_0$))}$ \hspace{0.355cm}\%\,Instances in high dynamic scene.
		
		\For{each instance $b^i$ in ${\bB_0}$}{}		                                \vspace{0.5mm}
			\If{$\frac{\text{Length}({\b{ L}_0}[idx[b^i]]==1)}{\text{Length}[idx[b^i]]}>\alpha_0$}     \vspace{0.5mm}
				\State $\bI_0[b^i]=1$
			\EndIf
			\If{$\text{Instance\_label}(b^i)\hspace{-1mm}==\hspace{-1mm}car$}                      \vspace{0.5mm}
				\If{$\frac{\text{Length}({\b{ L}_0}[idx[b^i]]==1)}{\text{Length}[idx[b^i]]}>\alpha_1$} \vspace{0.5mm}
					\State $m_c = m_c +1$
				\EndIf                                                                      \vspace{0.5mm}
				\If{$\frac{\text{Length}({\b{ P}_0}[idx[b^i]]>\beta_0)}{\text{Length}[idx[b^i]]}>0.5$} \vspace{0.5mm}
					\State $ \bM[b^i]=1 $
				\EndIf
			\EndIf
		\EndFor
		
		\vspace{1mm}
		\For{each instance $b^i$ in ${\bB_0}$} \hspace{0.305cm}\% For high dynamic scene
			\If{$m_c>\beta_1$ and $ \bM[b^i]==1$}
				\State $\bI_0[b^i]=1$;
			\EndIf
		\EndFor
		
		\vspace{1mm}
		\State $\bC_B^0$, $\bO_B^0=\text{Get\_box}(\bB_0)$ \hspace{0.1cm}\%Get center and box of instances.
		\For{$j=1,...,\theta_{0}$} \hspace{0.805cm} \%Integrating past observations.
			\State $\bC_B^{0\rightarrow j} = \text{Transform}(\bC_B^0,\bT_0^{j})$\vspace{0.5mm}
			\State $\bC_B^j$, $\bO_B^j= \text{Get\_box}(\bB_j)$    \vspace{0.5mm}
			\If{$\text{Match}(\bC_B^{0\rightarrow j},\bC_B^j,\bO_B^0,\bO_B^j)$ and $\bI_j[\bB_j]==1$} \vspace{0.5mm}
				\State $\bQ[\bB^0] = \bQ[\bB^0]+1 $
			\EndIf 
		\EndFor
		\State $\bI_0[\bQ[\bB^0]>\theta_{1}]=1$
		
		\vspace{1mm}
		\For{each instance $b^i$ in ${\b{ B}_0}$} \hspace{0.305cm}\%\,Output point-wise labels.
			\If{$\bI_0[b^i]==1$}
				\State ${\b{ L}_0}'[idx[b^i]]=1$
			\EndIf
		\EndFor
	\end{algorithmic}
\end{algorithm}

This algorithm aims to refine the results of the following three cases.
Firstly, if the percentage of the moving points in an instance is greater than $\alpha_0$, the instance is considered to be moving, and all the points of the instance will be labeled as moving (see lines 8-9). 
Secondly, if the number of moving vehicles in the scan exceeds $\beta_1$, the scene is considered to be highly dynamic, such as the highway, 
and all the points of the vehicle can be labeled as moving with lower confidence $\beta_0$, which means they can be easily labeled as moving (see lines 10-17).
Thirdly, if an instance has been classified as moving in the past $\theta_0$ scans for $\theta_1$ times, the instance is considered to be moving in the current scan as well (see lines 18-24).
Note that the refinement algorithm is only performed on the points with instance labels.
By this, we exploit the instance information in both top-down and bottom-up ways to improve the MOS performance.

\subsection{Loss and Network Training}
\label{sec:train}

We train our network once to achieve both instance detection and MOS. Our loss function is composed of motion loss~$L_\text{motion}$, instance classification loss~$L_\text{cls}$, bounding box regression loss~$L_\text{reg}$, and moving segmentation loss~$L_\text{mos}$:
\begin{equation}
	L_\text{total} = L_\text{motion}+L_\text{cls}+L_\text{reg}+L_\text{mos}.
\end{equation}

$L_\text{motion}$ and $L_\text{mos}$ use the CrossEntropy Loss function \cite{sun2022iros,mersch2022ral} defined as follow: 

\begin{equation}
	L_\text{motion}(y,\hat{y}) = -\sum_{i=1}^{M} \sum_{k=1}^{D}P_k(y_i)\text{log}(P_k(\hat{y}_i)),
\end{equation} 
\begin{equation}
	L_\text{mos}(y,\hat{y}') = -\sum_{i=1}^{M} \sum_{k=1}^{D}P_k(y_i)\text{log}(P_k(\hat{y}_i')),
\end{equation} 
where $y_i$ is the ground truth, $\hat{y}_i$ is the prediction of MotionNet and $\hat{y}_i'$ is the prediction of upsample fusion module. $P_k(y)$ denotes probability that $y$ belongs to the $k$-th class in $[\text{unlabeled},\text{static},\text{moving}]$, and  $P_k(\hat{y}_i)=\text{softmax}(\b{f}_i)$, $P_k(\hat{y}_i')=\text{softmax}(\b{f}'_i)$.

We follow \cite{law2018eccv} to use a variant focal loss for $L_\text{cls}$ and a smooth L1 loss for $L_\text{reg}$:
\begin{equation}
\footnotesize
L_\text{cls} \hspace{-1mm}=\hspace{-0.8mm}\frac{-1}{U}\hspace{-1mm}\sum_{c=1}^{C}\hspace{-0.5mm}\sum_{i=1}^{H}\hspace{-0.5mm}\sum_{j=1}^{W}\hspace{-0.8mm}\begin{cases}
			\hspace{5mm}(1-\hat{E}_{cij})^{\sigma}{\rm log}(\hat{E}_{cij})& \hspace{-2mm}{\rm if}\hspace{2mm} \hspace{-1mm}g_{cij}\hspace{-0.6mm}=\hspace{-0.6mm}1\\
			\hspace{-0.5mm}(1\hspace{-0.8mm}-\hspace{-0.8mm}g_{cij})^{\gamma}(\hat{E}_{cij})^{\sigma}{\rm log}(1\hspace{-0.8mm}-\hspace{-0.8mm}\hat{E}_{cij})& \hspace{-2mm}{\rm otherwise}
															\end{cases},
\end{equation}
\begin{equation}
L_\text{reg} = \frac{1}{U}\sum_{u=1}^{U}\text{smooth-L1-loss}(\b{o}_u,\b{\hat{o}}_u), 
\end{equation}
where $g_{cij}$ indicates the Gaussian bumps of the heatmap encoded through the ground truth, both $\sigma$ and $\gamma$ are the hyper parameters, and $\b{o}_u$ represents the ground truth.
We refer more details for $L_\text{cls}$ and $L_\text{reg}$ in \cite{yin2021cvpr,law2018eccv}.

\section{Experimental Evaluation}
\label{sec:exp}

%
%
We conduct the experiments to show that our approach
(i) predicts instances and segments moving objects simultaneously;
(ii) achieves the state-of-the-art MOS performance compared with existing methods;
and 
(iii) generalizes well to different scenes without any parameter adaptation.

\subsection{Experimental Setup}
\textbf{Datasets.} We train our model on both SemantiKITTI-MOS dataset \cite{chen2021ral,behley2019iccv} and KITTI-road dataset\footnote{https://www.cvlibs.net/datasets/kitti/raw\_data.php?type=road} and evaluate it on the SemantiKITTI-MOS benchmark\footnote{https://codalab.lisn.upsaclay.fr/competitions/7088}.
SemantiKITTI-MOS dataset contains a total of 22 sequences with a point-wise semantic category of moving or static. We use sequences 00-07 and 09-10 for training, sequence 08 for validation, and sequences 11-21 for testing.
To overcome the imbalance distribution of quantities between moving and static objects on the SemantiKITTI-MOS dataset, we follow~\cite{sun2022iros} and use their conducted training and validation sets based on KITTI-road, where sequences 30-34, 40 for training and 35-39, 41 for validation. 

In addition to the ground truth of moving objects, our method also requires instance information for training. We generate 3D bounding boxes for both the SemanticKITTI dataset and KITTI-road dataset by first enclosing the clusters generated by the Euclidean clustering algorithm in PCL library~\cite{rusu2011icra}, and then manually modifying erroneous cases. We will release the instance data together with our code for the convenience of the community.
We furthermore show our instance detection results on the KITTI-tracking dataset\footnote{https://www.cvlibs.net/datasets/kitti/eval\_tracking\_overview.php}. 

\textbf{Evaluation Metrics.} We use intersection-over-union (IoU) \cite{everingham2010ijcv} of the moving objects as the evaluation metric:
\begin{equation}
\text{IoU} = \frac{\text{TP}}{\text{TP}+\text{FP}+\text{FN}},
\end{equation}
where TP, FP, and FN represent true positive, false positive and false negative predictions of moving points.

\textbf{Training Details.} We set the number of consecutive input scans to $N=10$. Before being taken as input, the point clouds are quantized into 4D voxels with a time resolution $\Delta t=0.1$\,s and spatial resolution $\Delta s=0.1\times0.1\times0.1$\,m.
In the refinement, we set $\alpha_0=0.6$, $\alpha_1=0.3$, $\beta_0=10^{-5}$, $\beta_1=5$, $\theta_{0}=5$ and $\theta_{1}=3$.  

Our framework is implemented in PyTorch~\cite{paszke2019neurips} and trained on 4 NVIDIA RTX 3090 GPUs.  
We train the whole model using the Adam optimizer~\cite{kingma2014iclr} with a batch size of 16. The learning rate is initialized as $10^{-4}$ and decays exponentially by 0.01 every epoch. We train the network for total 160 epochs to achieve the best performance. And the widely-used data-augmentation strategies: random flipping, scaling, and rotations are adopted to boost the performance.
\begin{table}[t]
	\caption{Performance comparison with other baseline methods on SemanticKITTI-MOS benchmark. Best result in bold.}
	\centering
	\begin{threeparttable}
	\begin{tabular}{m{6.5cm} cm{8mm}}
		\toprule
		&      IoU[\%]   \\
		\midrule
		SpSequenceNet & $43.2$ \\
		KPConv   & $60.9$ \\
		\midrule
		AutoMOS$^*$    & $62.3$ \\
		LMNet$^\dagger$    & $62.5$ \\
		4DMOS   &  $65.2$\\
		MotionSeg3D$^*$   & $70.2$ \\
		RVMOS$^\dagger$   & $74.7$ \\
		\midrule
		Ours$^*$ \footnotesize{($N=10$~Scans, $\Delta t=0.1$~s)}    & $\bf{75.6}$ \\
		\bottomrule
	\end{tabular}
	\begin{tablenotes}
	\footnotesize
	\item $^*$ indicates the method is trained both on the SemanticKITTI dataset and KITTI-road dataset.
	\item $^\dagger$ indicates the method exploiting semantic labels.
	\end{tablenotes}

	\end{threeparttable}
	\label{tab:performance_test}
\end{table}

\begin{figure*}[ht]
	\centering
	\includegraphics[width=1\linewidth]{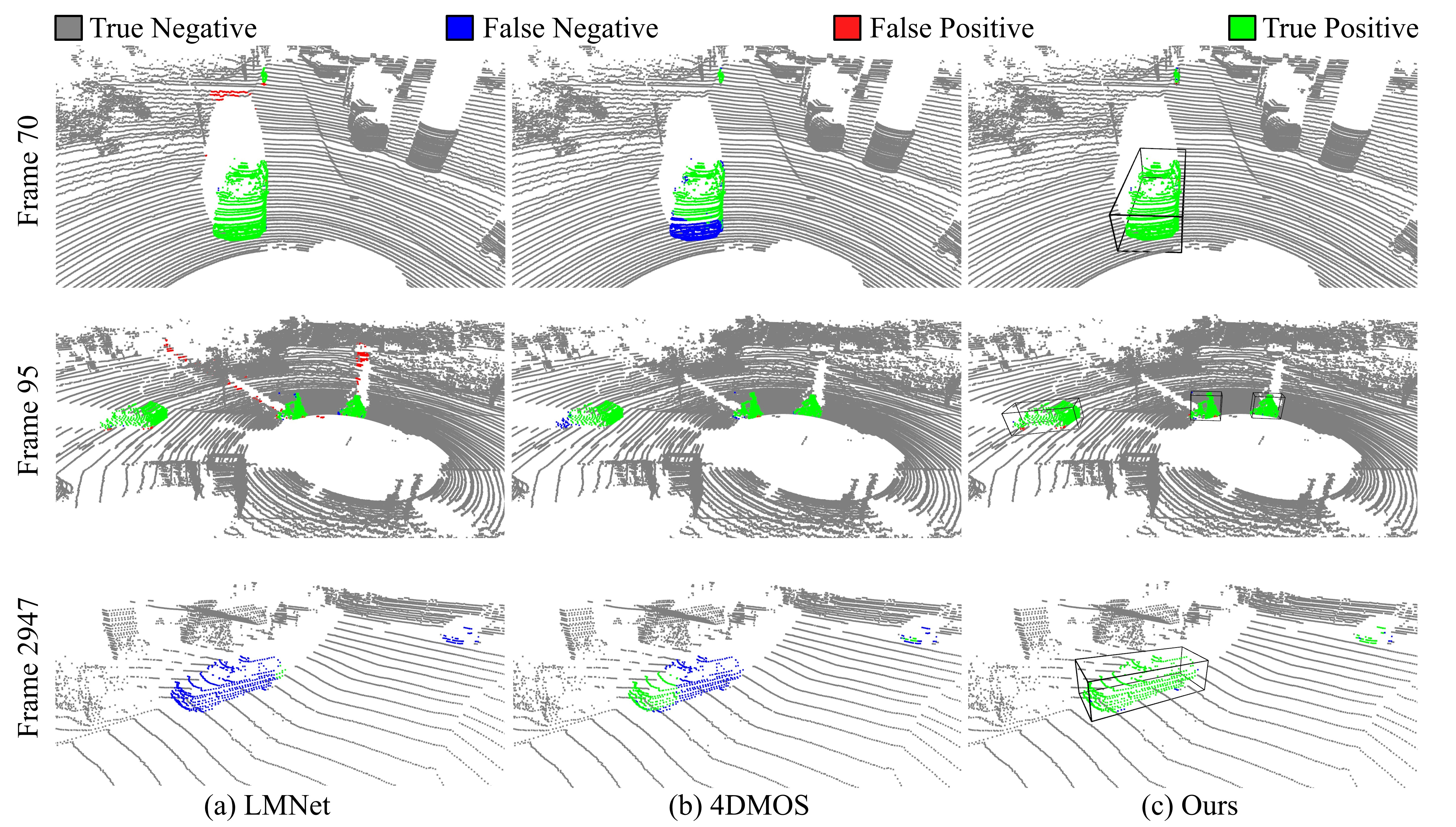}
	\caption{Qualitative results of MOS compared with LMNet and 4DMOS on SemanticKITTI validation set. Best viewed in color.
	}
	\label{fig:qualitative}
\end{figure*}

\subsection{MOS Performance Evaluation}
We evaluate the performance of our method in SemanticKITTI-MOS benchmark and compare it with several baseline methods, including  SpSequenceNet~\cite{shi2020cvpr}, KPConv~\cite{thomas2019iccv}, AutoMos~\cite{chen2022ral}, LMNet~\cite{chen2021ral}, 4DMOS~\cite{mersch2022ral}, MotionSeg3D~\cite{sun2022iros} and RVMOS~\cite{kim2022ral}.
SpSequenceNet and KPConv are originally used for semantic segmentation, while we present their MOS results reported in~\cite{chen2022ral}. 
In addition, we obtain the experimental results of all other methods from the SemanticKITTI-MOS benchmark, which are also made public in their paper.
 
The quantitative comparison is shown in~\tabref{tab:performance_test}, and the results support our second claim that our approach achieves the state-of-the-art MOS performance with $75.6\%$ IoU.
LMNet, 4DMOS and RVMOS are trained on the SemantiKITTI dataset, and RVMOS achieves great performance improvement up to $74.7\%$ IoU. That is mainly due to the fact that RVMOS inserts semantic information in the network implementations.
AutoMos, MotionSeg3D and our network perform training on the SemantiKITTI and an additional labeled KITTI road dataset~\cite{chen2021ral,sun2022iros} in order to reduce the impact of unbalanced data distribution.

For a fair comparison, we also report the validation set result of our network trained only on the SemanticKITTI dataset and compare it with these baselines.
The 4DMOS result is re-evaluated using the same pose as ours, which is estimated by the SuMa~\cite{behley2018rss} algorithm with loop closure.    
As shown in \tabref{tab:performance_validation}, our approach still achieves the best performance.

\figref{fig:qualitative} visualizes qualitative comparisons of our method with LMNet and 4DMOS on the SemanticKITTI validation set. LMNet is the first work on LiDAR MOS exploiting range images, which is fast but results in many wrong predictions. 4DMOS performs well on fast-moving object segmentation, but not as well on slow-moving object segmentation. As 4DMOS cannot capture the instance information of the moving points, only partial points of the moving instance can be correctly predicted. 
However, our approach can segment moving objects completely and has the ability to detect slow-moving instances by integrating past observations in the instance-based refinement algorithm.
The qualitative experimental results support our first claim and further demonstrate that instance information is highly valuable for MOS task.

\begin{table}[t]
	\caption{Performance comparison with other baseline methods on SemanticKITTI validation set, and all methods are trained using the same dataset. Best result in bold.}
	\centering
	\begin{tabular}{m{6.5cm} cm{8mm}}
		\toprule
		&      IoU[\%]   \\
		\midrule
		LiMoSeg & $52.6$ \\
		LMNet    & $67.1$ \\
		MotionSeg3D   & $68.1$ \\
		RVMOS   & $71.2$ \\
		4DMOS   &  $71.9$\\
		\midrule
		Ours \footnotesize{($N=10$~Scans, $\Delta t=0.1$~s)}   & $\bf{73.2}$ \\
		\bottomrule
	\end{tabular}
	\label{tab:performance_validation}
\end{table}

\begin{table}[t]
	\caption{The generalization performance evaluation on the Apollo dataset. Best result in bold.}
	\centering
	\begin{tabular}{m{6.5cm} cm{8mm}}
		\toprule
		&      IoU[\%]   \\
		\midrule
		MotionSeg3D & $7.5$ \\
		LMNet   & $16.9$ \\
		LMNet+AutoMOS+Fine-Tuned   & $65.9$ \\
		4DMOS &  $73.1$\\
		\midrule
		Ours \footnotesize{($N=10$~Scans, $\Delta t=0.1$~s)}  & $\bf{78.0}$ \\
		\bottomrule
	\end{tabular}
	\vspace{-0.4cm}
	
	\label{tab:generalization}
\end{table}

\begin{table*}[t]
	\caption{Ablation study of different components and setups in SemanticKITTI validation set. $N$ and $\Delta t$ refer to the number and the temporal resolution of input sequence LiDAR scans, mentioned in \secref{sec:pre}. The instance layer is the component of the Instance Pyramid. Refinement refers to our proposed instance-based refinement algorithm in \secref{sec:post_processing}.}
	\centering
	\begin{tabular}{p{0.3cm}|p{11.2cm}|p{0.5cm}<{\centering}|p{0.5cm}<{\centering}|p{0.5cm}<{\centering}|p{0.5cm}<{\centering}|p{1cm}<{\centering}}
		\toprule
		& \multirow{2}{*}{\textbf{Method}}  & \multicolumn{2}{c|}{\textbf{Train}} &\multicolumn{2}{c|}{\textbf{Inference}}    \\
		&   & $N$ & $\Delta t$ &$N$       &$\Delta t$                      & IoU[\%]    \\
		\midrule
		$[$A$]$  & MotionNet & $5$ & $0.1$ & $5$ & $0.1$ & $55.6$\\
		$[$B$]$  & + Instance Detection + Upsample Fusion(without instance) & $5$ & $0.1$ & $5$ & $0.1$ & $59.8$\\
		$[$C$]$  & + Instance Detection + Upsample Fusion(with 1 instance layer) & $5$ & $0.1$ & $5$ & $0.1$ & $60.4$\\
		$[$D$]$  & + Instance Detection + Upsample Fusion(with 4 instance layers) & $5$ & $0.1$ & $5$ & $0.1$ & $60.8$\\
		$[$E$]$  & + Instance Detection + Upsample Fusion(with 4 instance layers) & $5$ & $0.2$ & $5$ & $0.2$ & $63.2$\\
		$[$F$]$  & + Instance Detection + Upsample Fusion(with 4 instance layers) & $10$ & $0.1$ & $5$ & $0.1$ & $65.2$\\
		$[$G$]$  & + Instance Detection + Upsample Fusion(with 4 instance layers) & $10$ & $0.1$ & $10$ & $0.1$ & $70.7$\\
		$[$H$]$  & + Instance Detection + Upsample Fusion(with 4 instance layers) + Refinement & $10$ & $0.1$ & $10$ & $0.1$ & $73.2$\\
		\bottomrule
	\end{tabular}
	\vspace{-0.3cm}
	
	\label{tab:ablation}
\end{table*}

\subsection{Generalization Analyses}
To demonstrate the third claim that our approach generalizes well to different scenes, we conduct experiments on the Apollo~\cite{lu2019cvpr} dataset.
We compare our method with three open source baseline methods, including MotionSeg3D~\cite{sun2022iros}, LMNet \cite{chen2021ral} and 4DMOS \cite{mersch2022ral}. 
All methods are only trained on the training set of SemanticKITTI and evaluated on the Apollo dataset without modifying any settings or parameters fine-tuning.
Besides, we also present the result of fine-tuned LMNet combined with AutoMOS~\cite{chen2022ral}, noted as LMNet+AutoMOS+Fine-Tuned~\cite{chen2021ral}, see \cite{chen2022ral} for details.
The results are shown in \tabref{tab:generalization}. Two range image-based methods MotionSeg3D and LMNet show bad performance in the generalization test, while 4DMOS and our method maintain good segmentation ability in unknown environments.
The possible reason is that the projection-based approaches overfit the setup of the sensor and specific patterns in the trained environments.
This may cause the network always to assume that the moving objects are found in a specific area of range images, which will affect the performance of projection-based approaches while the point cloud-based approaches will not be affected.
Benefiting from the learned instance information, our method outperforms all baseline methods. 

\subsection{Ablation Study}
\label{sec:Ablation}
We conduct ablation studies on the SemanticKITTI validation set to better understand the effectiveness of individual modules in our method.
We train all models for 80 epochs and report the best results, shown in \tabref{tab:ablation}.

From the comparison of $[$A$]$ and $[$B$]$, we show that the performance is boosted by introducing the spatio-temporal information extracted by our instance detection module and upsample fusion module. By integrating the instance information, the results are further improved, as can be seen in the comparison of $[$B$]$ and $[$C$]$, $[$D$]$.
The advantage of introducing the instance information can be further demonstrated by comparing $[$G$]$ and $[$H$]$. 
 
The comparison of $[$D$]$, $[$F$]$ and $[$G$]$ shows that feeding more information to the network by increasing the number of input scans $N$ both in training (see $[$D$]$ and $[$F$]$) or the inference (see $[$F$]$ and $[$G$]$) improves the performance.
Raising temporal resolution $\Delta t$ can also improve the performance, as can be seen in the comparison of $[$D$]$ and $[$E$]$. 
The reason is increasing the temporal resolution will extend the time horizon of the input sequence, which can help the MotionNet capture the motion information.

\subsection{3D Instance Detection}
\begin{figure}[t]
	\centering
	\subfigcapskip=4pt
	\subfigure[Frame 158]{
		\hspace{-0.25cm}
		\begin{minipage}[t]{\linewidth}
			\centering
			\includegraphics[width=3.2in]{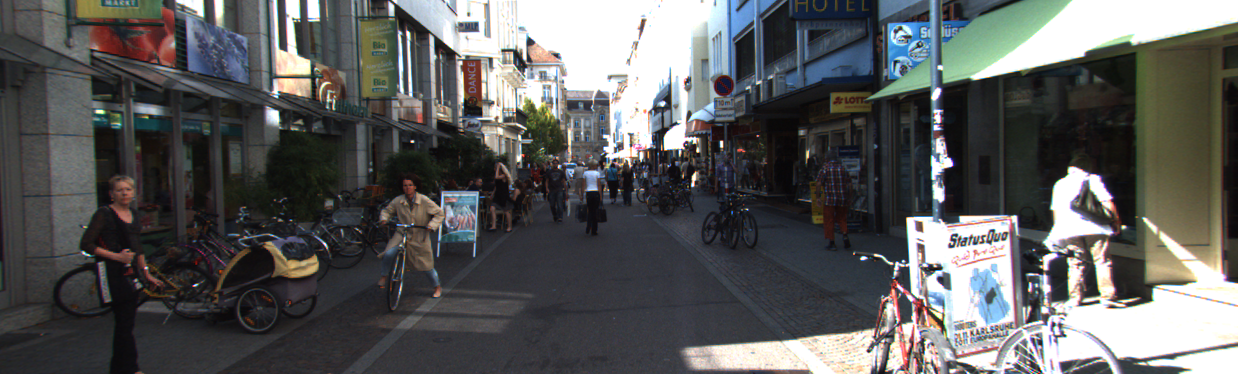}
			\includegraphics[width=3.2in]{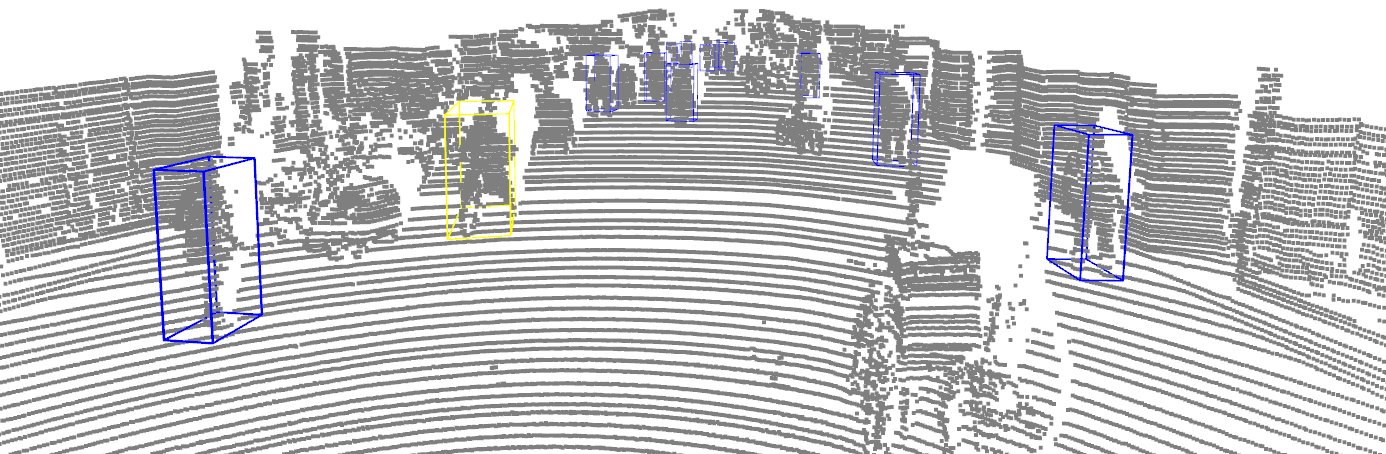}
		\end{minipage}%
	}%
	\quad
	\subfigure[Frame 674 ]{
		\hspace{-0.25cm}
		\begin{minipage}[t]{\linewidth}
			\centering
			\includegraphics[width=3.2in]{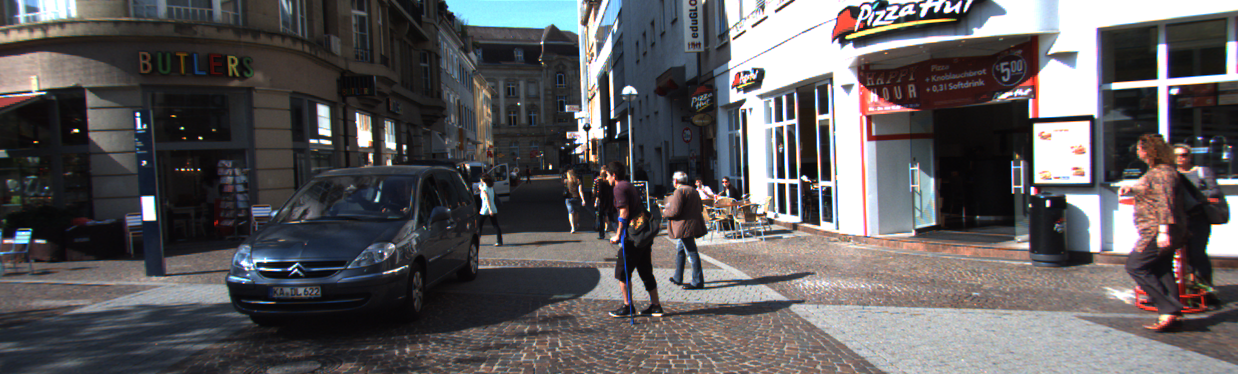}
			\includegraphics[width=3.2in]{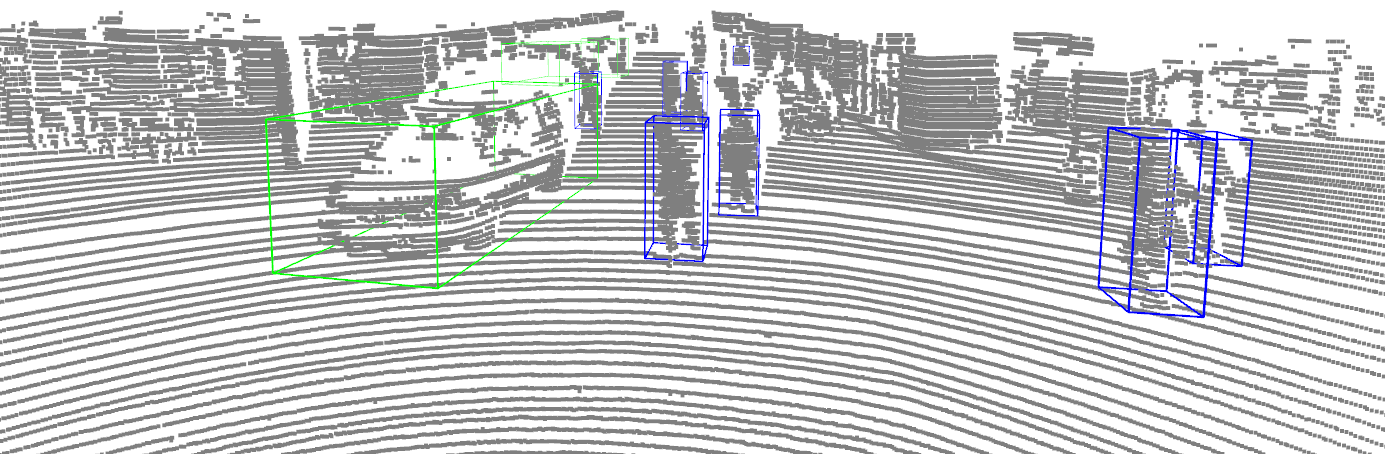}
		\end{minipage}%
	}%
	\centering
	\caption{The instance detection results of main traffic participants in sequence 19 of the KITTI tracking dataset. The blue bounding boxes represent pedestrians, the yellow bounding boxes represent cyclists, and the green bounding boxes represent cars.}
	\label{3Dinsatnce}
	\vspace{-0.3cm}
\end{figure}

To further demonstrate the ability of our method to detect instances, we evaluate the performance on the KITTI tracking dataset with the model pre-trained on SemanticKITTI. The KITTI tracking dataset contains more instances compared to the SemanticKITTI dataset.
The results presented in \figref{3Dinsatnce} show that our method is able to accurately predict the instance category and 3D bounding box of major traffic participants, albeit in a complex environment. 

\subsection{Runtime}
We measure the runtime of our unoptimized Python implementation method with a single NVIDIA RTX 3090 GPU. Our network takes an average time of 82.4~ms for $N=5$ and 120~ms for $N=10$. And refinement takes 7~ms.
According to the ablation studies in~\secref{sec:Ablation}, one can choose a suitable number of input scans $N$ to achieve faster inference with a little loss of accuracy.

\section{Conclusion}
\label{sec:conclusion}
In this paper, we presented a novel network to predict both point-wise moving labels and instance information in the current scan.
Our network extracts motion features, spatio-temporal features and instance information from different modules and fuses them to achieve moving object segmentation.
By again integrating instance information in the instance-based refinement algorithm, our approach can distinguish between moving and static instances and fully segment the points within moving instances.
The experimental results on the SemanticKITTI and the Apollo dataset demonstrate that our approach is effective and generalizes well to different environments.

\bibliographystyle{unsrt}
\bibliography{new,glorified}

\begin{thebibliography}{10}

\bibitem{kim2022ral}
J.~Kim, J.~Woo, and S.~Im.
\newblock Rvmos: Range-view moving object segmentation leveraged by semantic
  and motion features.
\newblock {\em IEEE Robotics and Automation Letters (RA-L)}, 7(3):8044--8051,
  2022.

\bibitem{lu2019iccv}
W.~Lu, G.~Wan, Y.~Zhou, X.~Fu, P.~Yuan, and S.~Song.
\newblock Deepvcp: An end-to-end deep neural network for point cloud
  registration.
\newblock In {\em Proc.~of the IEEE/CVF Intl.~Conf.~on Computer Vision (ICCV)},
  pages 12--21, 2019.

\bibitem{chen2019iros}
X.~Chen, A.~Milioto, E.~Palazzolo, P.~Giguère, J.~Behley, and C.~Stachniss.
\newblock {SuMa++: Efficient LiDAR-based Semantic SLAM}.
\newblock In {\em Proc.~of the IEEE/RSJ Intl.~Conf.~on Intelligent Robots and
  Systems (IROS)}, 2019.

\bibitem{chen2020iros}
X.~Chen, T.~L\"abe, L.~Nardi, J.~Behley, and C.~Stachniss.
\newblock {Learning an Overlap-based Observation Model for 3D LiDAR
  Localization}.
\newblock In {\em Proc.~of the IEEE/RSJ Intl.~Conf.~on Intelligent Robots and
  Systems (IROS)}, 2020.

\bibitem{arora2021ecmr}
M.~Arora, L.~Wiesmann, X.~Chen, and C.~Stachniss.
\newblock {Mapping the Static Parts of Dynamic Scenes from 3D LiDAR Point
  Clouds Exploiting Ground Segmentation}.
\newblock In {\em Proc.~of the Europ.~Conf.~on Mobile Robotics (ECMR)}, 2021.

\bibitem{kim2020iros}
Giseop Kim and Ayoung Kim.
\newblock Remove, then revert: Static point cloud map construction using
  multiresolution range images.
\newblock In {\em Proc.~of the IEEE/RSJ Intl.~Conf.~on Intelligent Robots and
  Systems (IROS)}, 2020.

\bibitem{chen2022neuroc}
P.~Chen, J.~Pei, W.~Lu, and M.~Li.
\newblock A deep reinforcement learning based method for real-time path
  planning and dynamic obstacle avoidance.
\newblock {\em Neurocomputing}, 497:64--75, 2022.

\bibitem{lee2023ral}
C.~Lee and K.~Song.
\newblock Path re-planning design of a cobot in a dynamic environment based on
  current obstacle configuration.
\newblock {\em IEEE Robotics and Automation Letters (RA-L)}, 2023.

\bibitem{chen2021ral}
X.~Chen, S.~Li, B.~Mersch, L.~Wiesmann, J.~Gall, J.~Behley, and C.~Stachniss.
\newblock {Moving Object Segmentation in 3D LiDAR Data: A Learning-based
  Approach Exploiting Sequential Data}.
\newblock {\em IEEE Robotics and Automation Letters (RA-L)}, 6:6529--6536,
  2021.

\bibitem{sun2022iros}
J.~Sun, Y.~Dai, X.~Zhang, J.~Xu, R.~Ai, W.~Gu, and X.~Chen.
\newblock Efficient spatial-temporal information fusion for lidar-based 3d
  moving object segmentation.
\newblock In {\em Proc.~of the IEEE/RSJ Intl.~Conf.~on Intelligent Robots and
  Systems (IROS)}, pages 11456--11463. IEEE, 2022.

\bibitem{mersch2022ral}
B.~Mersch, X.~Chen, I.~Vizzo, L.~Nunes, J.~Behley, and C.~Stachniss.
\newblock Receding moving object segmentation in 3d lidar data using sparse 4d
  convolutions.
\newblock {\em IEEE Robotics and Automation Letters (RA-L)}, 7(3):7503--7510,
  2022.

\bibitem{kreutz2023wacv}
T.~Kreutz, M.~M{\"u}hlh{\"a}user, and A.~S. Guinea.
\newblock Unsupervised 4d lidar moving object segmentation in stationary
  settings with multivariate occupancy time series.
\newblock In {\em Proc.~of the IEEE Winter Conf.~on Applications of Computer
  Vision (WACV)}, pages 1644--1653, 2023.

\bibitem{baur2021iccv}
Stefan~Andreas Baur, David~Josef Emmerichs, Frank Moosmann, Peter Pinggera,
  Bjorn Ommer, and Andreas Geiger.
\newblock {SLIM: Self-Supervised LiDAR Scene Flow and Motion Segmentation}.
\newblock In {\em Proc.~of the IEEE/CVF Intl.~Conf.~on Computer Vision (ICCV)},
  2021.

\bibitem{liu2019cvpr}
Xingyu Liu, Charles~R Qi, and Leonidas~J Guibas.
\newblock {FlowNet3D: Learning Scene Flow in 3D Point Clouds}.
\newblock In {\em Proc.~of the IEEE/CVF Conf.~on Computer Vision and Pattern
  Recognition (CVPR)}, 2019.

\bibitem{dong2022cvpr}
G.~Dong, Y.~Zhang, H.~Li, X.~Sun, and Z.~Xiong.
\newblock Exploiting rigidity constraints for lidar scene flow estimation.
\newblock In {\em Proc.~of the IEEE/CVF Conf.~on Computer Vision and Pattern
  Recognition (CVPR)}, pages 12776--12785, 2022.

\bibitem{pomerleau2014icra}
F.~Pomerleau, P.~Kr{\"u}siand, F.~Colas, P.~Furgale, and R.~Siegwart.
\newblock Long-term 3d map maintenance in dynamic environments.
\newblock In {\em Proc.~of the IEEE Intl.~Conf.~on Robotics \& Automation
  (ICRA)}, 2014.

\bibitem{pagad2020icra}
Shishir Pagad, Divya Agarwal, Sathya Narayanan, Kasturi Rangan, Hyungjin Kim,
  and Ganesh Yalla.
\newblock {Robust Method for Removing Dynamic Objects from Point Clouds}.
\newblock In {\em Proc.~of the IEEE Intl.~Conf.~on Robotics \& Automation
  (ICRA)}, 2020.

\bibitem{ghorai2022tits}
P.~Ghorai, A.~Eskandarian, Y.~K. Kim, and G.~Mehr.
\newblock State estimation and motion prediction of vehicles and vulnerable
  road users for cooperative autonomous driving: A survey.
\newblock {\em IEEE Trans.~on Intelligent Transportation Systems (ITS)},
  23(10):16983--17002, 2022.

\bibitem{choy2019cvpr}
C.~Choy, J.~Gwak, and S.~Savarese.
\newblock 4d spatio-temporal convnets: Minkowski convolutional neural networks.
\newblock In {\em Proc.~of the IEEE/CVF Conf.~on Computer Vision and Pattern
  Recognition (CVPR)}, 2019.

\bibitem{chen2022ral}
X.~Chen, B.~Mersch, L.~Nunes, R.~Marcuzzi, I.~Vizzo, J.~Behley, and
  C.~Stachniss.
\newblock Automatic labeling to generate training data for online lidar-based
  moving object segmentation.
\newblock {\em IEEE Robotics and Automation Letters (RA-L)}, 7(3):6107--6114,
  2022.

\bibitem{mohapatra2021arxiv}
S.~Mohapatra, M.~Hodaei, S.~Yogamani, S.~Milz, H.~Gotzig, M.~Simon, H.~Rashed,
  and P.~Maeder.
\newblock Limoseg: Real-time bird's eye view based lidar motion segmentation.
\newblock {\em arXiv preprint}, 2021.

\bibitem{wu2020cvpr}
Pengxiang Wu, Siheng Chen, and Dimitris~N. Metaxas.
\newblock {MotionNet: Joint Perception and Motion Prediction for Autonomous
  Driving Based on Bird’s Eye View Maps}.
\newblock In {\em Proc.~of the IEEE/CVF Conf.~on Computer Vision and Pattern
  Recognition (CVPR)}, 2020.

\bibitem{lim2021ral}
Hyungtae Lim, Sungwon Hwang, and Hyun Myung.
\newblock {ERASOR: Egocentric Ratio of Pseudo Occupancy-Based Dynamic Object
  Removal for Static 3D Point Cloud Map Building}.
\newblock {\em IEEE Robotics and Automation Letters (RA-L)}, 6(2):2272--2279,
  2021.

\bibitem{graham2018cvpr}
B.~Graham, M.~Engelcke, and L.~van~der Maaten.
\newblock {3D Semantic Segmentation with Submanifold Sparse Convolutional
  Networks}.
\newblock In {\em Proc.~of the IEEE Conf.~on Computer Vision and Pattern
  Recognition (CVPR)}, 2018.

\bibitem{yin2021cvpr}
T.~Yin, X.~Zhou, and P.~Krahenbuhl.
\newblock {Center-Based 3D Object Detection and Tracking}.
\newblock In {\em Proc.~of the IEEE/CVF Conf.~on Computer Vision and Pattern
  Recognition (CVPR)}, 2021.

\bibitem{law2018eccv}
H.~Law and J.~Deng.
\newblock {CornerNet: Detecting Objects as Paired Keypoints}.
\newblock In {\em Proc.~of the Europ.~Conf.~on Computer Vision (ECCV)}, 2018.

\bibitem{behley2019iccv}
J.~Behley, M.~Garbade, A.~Milioto, J.~Quenzel, S.~Behnke, C.~Stachniss, and
  J.~Gall.
\newblock {SemanticKITTI: A Dataset for Semantic Scene Understanding of LiDAR
  Sequences}.
\newblock In {\em Proc.~of the IEEE/CVF Intl.~Conf.~on Computer Vision (ICCV)},
  2019.

\bibitem{rusu2011icra}
R.~B. Rusu and S.~Cousins.
\newblock 3d is here: Point cloud library (pcl).
\newblock In {\em Proc.~of the IEEE Intl.~Conf.~on Robotics \& Automation
  (ICRA)}, pages 1--4. IEEE, 2011.

\bibitem{everingham2010ijcv}
M.~Everingham, L.~Van~Gool, C.K. Williams, J.~Winn, and A.~Zisserman.
\newblock {The Pascal Visual Object Classes (VOC) Challenge}.
\newblock {\em Intl.~Journal~of Computer Vision (IJCV)}, 88(2):303--338, 2010.

\bibitem{paszke2019neurips}
A.~Paszke, S.~Gross, F.~Massa, A.~Lerer, J.~Bradbury, G.~Chanan, T.~Killeen,
  Z.~Lin, Natalia N.~Gimelshein, L.~Antiga, et~al.
\newblock {PyTorch: An Imperative Style, High-Performance Deep Learning
  Library}.
\newblock In {\em Proc.~of the Conference on Neural Information Processing
  Systems (NeurIPS)}, pages 8026--8037, 2019.

\bibitem{kingma2014iclr}
D.~Kingma and J.~Ba.
\newblock Adam: A method for stochastic optimization.
\newblock {\em Proc.~of the Int.~Conf.~on Learning Representations (ICLR)},
  abs/1412.6980, 2015.

\bibitem{shi2020cvpr}
Hanyu Shi, Guosheng Lin, Hao Wang, Tzu-Yi Hung, and Zhenhua Wang.
\newblock {SpSequenceNet: Semantic Segmentation Network on 4D Point Clouds}.
\newblock In {\em Proc.~of the IEEE/CVF Conf.~on Computer Vision and Pattern
  Recognition (CVPR)}, 2020.

\bibitem{thomas2019iccv}
H.~Thomas, C.R. Qi, J.~Deschaud, B.~Marcotegui, F.~Goulette, and L.J. Guibas.
\newblock {KPConv: Flexible and Deformable Convolution for Point Clouds}.
\newblock In {\em Proc.~of the IEEE/CVF Intl.~Conf.~on Computer Vision (ICCV)},
  2019.

\bibitem{behley2018rss}
J.~Behley and C.~Stachniss.
\newblock {Efficient Surfel-Based SLAM using 3D Laser Range Data in Urban
  Environments}.
\newblock In {\em Proc.~of Robotics: Science and Systems (RSS)}, 2018.

\bibitem{lu2019cvpr}
W.~Lu, Y.~Zhou, G.~Wan, S.~Hou, and S.~Song.
\newblock {L3-Net: Towards Learning Based LiDAR Localization for Autonomous
  Driving}.
\newblock In {\em Proc.~of the IEEE/CVF Conf.~on Computer Vision and Pattern
  Recognition (CVPR)}, 2019.

\end{thebibliography}
\end{document}